\documentclass[journal,twoside,web]{ieeecolor}
\usepackage{generic}
\usepackage{url}
\usepackage{subcaption}
\usepackage{pifont}
\usepackage{threeparttable}
\usepackage{amsmath,amssymb,amsfonts}
\usepackage{algorithmic}
\usepackage{graphicx}
\usepackage{multirow}
\usepackage{algorithm,algorithmic}
\usepackage{xcolor}
\definecolor{darkred}{rgb}{0.55, 0.0, 0.0}
\definecolor{darkspringgreen}{rgb}{0.09, 0.45, 0.27}
\definecolor{steelblue}{rgb}{0.27, 0.51, 0.71}

\newcommand{\blue}[1]{\textcolor{blue}{#1}}
\newcommand{\red}[1]{\textcolor{red}{#1}}

\newcommand{\darkspringgreen}[1]{\textcolor{darkspringgreen}{#1}}

\usepackage{cite}
\usepackage{hyperref} 
\hypersetup{
  colorlinks=true,       
  linkcolor=blue,        
  citecolor=blue,        
  filecolor=magenta,     
  urlcolor=blue
}
\usepackage[switch]{lineno}

\newcommand{\etal}[1]{#1~\textit{et al.}}

\usepackage{textcomp}
\usepackage{utfsym}
\newcommand{\newcheckmark}{\usym{1F5F8}}

\def\BibTeX{{\rm B\kern-.05em{\sc i\kern-.025em b}\kern-.08em
    T\kern-.1667em\lower.7ex\hbox{E}\kern-.125emX}}
\usepackage{atbegshi}

\newcommand{\topnotice}{%
   \begin{minipage}{\dimexpr\paperwidth-2cm} 
        \centering
        \scriptsize 
        This article has been accepted for publication in IEEE Journal of Biomedical and Health Informatics. 
        This is the author's version which has not been fully edited and content may change prior to final publication. 
        Citation information: DOI 10.1109/JBHI.2025.3613010
    \end{minipage}%
}

\AtBeginShipout{%
    \ifnum\value{page}=1
        \AtBeginShipoutUpperLeft{%
            \raisebox{-1.5\baselineskip}[0pt][0pt]{\makebox[\paperwidth][c]{\topnotice}}%
        }%
    \fi
}

\begin{document}

\title{LatXGen: Towards Radiation-Free and Accurate Quantitative Analysis of Sagittal Spinal Alignment Via Cross-Modal Radiographic View Synthesis}

\author{Moxin Zhao\textsuperscript{\dag},
Nan Meng\textsuperscript{\dag}$^{*}$,~\IEEEmembership{Member, IEEE},
Jason Pui Yin Cheung$^{*}$,
Chris Yuk Kwan Tang,
Chenxi Yu,
Wenting Zhong,
Pengyu Lu,
Chang Shi,
Yipeng Zhuang,
Teng Zhang$^{*}$, \IEEEmembership{Senior Member, IEEE}\vspace*{-2em}
\thanks{\textsuperscript{\dag} M.~Zhao and N.~Meng contributed equally to this work as co-first authors.}
\thanks{$^{*}$ Corresponding authors: N.~Meng, J.P.Y. Cheung and T.~Zhang}%
\thanks{M.~Zhao, N.~Meng, J.P.Y.~Cheung, C.Y.K.~Tang, W.~Zhong, P.~Lu, S.~Chang, Y.~Zhuang, and T.~Zhang are with the Department of Orthopaedics and Traumatology, The University of Hong Kong, Hong Kong SAR, China (e-mail: moxin@connect.hku.hk, nanmeng@hku.hk, cheungjp@hku.hk, medic.chris.tang@gmail.com, wtzhong@hku.hk, pengyuloo@connect.hku.hk, chaseshi@hku.hk, yipengzh@hku.hk, tgzhang@hku.hk).}
\thanks{C.~Yu is with the Department of Joint Surgery, Shandong Provincial Hospital Affiliated to Shandong First Medical University, Jinan, Shandong, China (e-mail: nameychxbj@gmail.com).}
\thanks{This work is supported in part by National Natural Science Foundation of China Young Scientists Fund (ID: 82303957 and ID: 82402398),  Innovation and Technology Fund (MRP/038/20X) and Health and Medical Research Fund (HMRF) 08192266}
}

\maketitle

\begin{abstract}
Adolescent Idiopathic Scoliosis (AIS) is a complex three-dimensional spinal deformity, and accurate morphological assessment requires evaluating both coronal and sagittal alignment. While previous research has made significant progress in developing radiation-free methods for coronal plane assessment, reliable and accurate evaluation of sagittal alignment without ionizing radiation remains largely underexplored.
To address this gap, we propose LatXGen, a novel generative framework that synthesizes realistic lateral spinal radiographs from posterior Red-Green-Blue and Depth (RGBD) images of unclothed backs. This enables accurate, radiation-free estimation of sagittal spinal alignment. LatXGen tackles two core challenges: (1) inferring sagittal spinal morphology changes from a lateral perspective based on posteroanterior surface geometry, and (2) performing cross-modality translation from RGBD input to the radiographic domain. The framework adopts a dual-stage architecture that progressively estimates lateral spinal structure and synthesizes corresponding radiographs. To enhance anatomical consistency, we introduce an attention-based Fast Fourier Convolution (FFC) module for integrating anatomical features from RGBD images and 3D landmarks, and a Spatial Deformation Network (SDN) to model morphological variations in the lateral view. Additionally, we construct the first large-scale paired dataset for this task, comprising 3,264 RGBD and lateral radiograph pairs. Experimental results demonstrate that LatXGen produces anatomically accurate radiographs and outperforms existing GAN-based methods in both visual fidelity and quantitative metrics. This study offers a promising, radiation-free solution for sagittal spine assessment and advances comprehensive AIS evaluation.

\end{abstract}


\section{Introduction}
\label{sec:introduction}
\IEEEPARstart{A}{dolescent} Idiopathic Scoliosis (AIS) is a three-dimensional (3D) spinal deformity affecting 2–4\% of adolescents aged 10–18~\cite{Fong2015Population}.It is characterized by lateral deviation of the spine (Cobb angle $\ge 10^\circ$), often accompanied by vertebral rotation and abnormal sagittal curvatures~\cite{Mak2021Patterns}. Although often asymptomatic in early stages, AIS can rapidly progress during growth spurts, leading to trunk deformity and, in severe cases with large curves, respiratory compromise if left untreated~\cite{Cheung2018Curve}. Early screening and continuous monitoring are therefore essential for timely intervention~\cite{Weinstein2013Effects}.

The clinical gold standard for AIS assessment is Cobb angle measurement from posteroanterior (PA) radiographs~\cite{Chung2018Spinal}. However, 
repeated radiographic examinations raise serious health concerns of radiation exposure in adolescents~\cite{Knott2014Sosort}.
To mitigate this, radiation-free alternatives such as forward-bending tests, scoliometers measurements, and Moir\'e topography have been explored~\cite{Cote1998Study,Choi2017CNN}, 
but they often lack accuracy and reliability due to subjectivity, inter-observer variability, and measurement inconsistencies, limiting their role in replacing radiographic assessments~\cite{Zhang2023Deep}.

Recent advances in optical imaging, particularly depth-sensing technologies such as Light Detection and Ranging (LiDAR) and structured-light Red-Green-Blue and Depth (RGBD) cameras, have enabled precise, radiation-free spinal evaluations~\cite{Roriz2021Automotive}. These technologies 
capture high-resolution back-surface geometry for AIS screening and diagnosis~\cite{Potts2024Lidar}. 
Prior studies used RGBD sensors to scan patients' back surfaces and predict Cobb angles from 3D point clouds~\cite{Kokabu2021Algorithm}. Some of these methods require subjects to maintain challenging postures, such as forward bending, leading to variability in image acquisition. Compared to forward-bending positions, 
standing-posture imaging is more feasible and has been used more for curvature assessment. 
\etal{Xu}~\cite{Xu2020Back} and \etal{Seoud}~\cite{Seoud2010Prediction} leveraged standing-posture 3D imaging for spinal curvature assessment; however, their methods depend heavily on manual computation, introducing significant inter-observer variability.

The advent of deep learning has further advanced optical imaging-based spinal analysis, enabling reliable clinical applications.
\etal{Meng}~\cite{Meng2023Radiograph} proposed a generative model using posterior RGBD data to synthesize coronal spinal radiographs, allowing radiation-free scoliosis quantification.
Later studies reduced dependence on depth data, showing that coronal-plane radiographs can even be generated from RGB images alone~\cite{He2024Conditional}.
Nevertheless, these methods mainly address coronal deformities. In clinical practice, coronal and sagittal deformities are often closely interrelated. 
Since coronal and sagittal deformities are closely related~\cite{Mak2021Patterns}, focusing only on the coronal plane is insufficient.
Reliable inference of sagittal spinal alignment parameters from optical back surface images thus remains a critical yet unresolved challenge.

To bridge this research gap, in this study, we aim to tackle a challenging yet clinically meaningful task: \textit{synthesizing realistic lateral spinal X-ray images directly from posteroanterior RGBD unclothed back images}, enabling radiation-free quantitative analysis of sagittal spinal alignment. This task inherently involves two major challenges: 1) accurate prediction of spinal morphological changes from a lateral perspective, and 2) cross-modal mapping from optical (RGBD) data to radiographic projection domains. To address these challenges, we propose a novel \textbf{Lat}eral \textbf{X}-ray image \textbf{Gen}eration (termed \textbf{LatXGen}) framework, which progressively estimates lateral spinal morphology and synthesizes corresponding lateral spinal X-ray images.
The main contributions of this study are summarized as follows:
\begin{itemize}
    \item We propose a novel dual-stage generative framework (LatXGen), which effectively integrates geometric features derived from posterior back surfaces and 3D anatomical landmarks to progressively predict lateral spinal curvature and synthesize realistic lateral spinal X-ray images. To the best of our knowledge, this work represents the first attempt to resolve \textit{cross-modal radiographic view synthesis} problem.
    \item We propose a novel attention-based fast Fourier convolution (FFC) layer to effectively integrate anatomical knowledge from both landmarks and RGBD images, along with a spatial deformation network (SDN) designed to accurately model spinal morphological variations in the lateral view.
    \item We establish the first large-scale cross-modal view synthesis dataset consisting of 3,264 paired RGBD back surface images and corresponding lateral spinal X-ray images. Experimental results demonstrate the superior performance of our proposed LatXGen model on lateral radiographic view synthesis compared to existing advanced approaches.
\end{itemize}

\section{Related Work}

\subsection{Cross-Modality Medical Image Generation}
Cross-modality image generation has recently attracted considerable interest, primarily driven by the need to augment datasets, reduce redundant imaging procedures, and enhance clinical efficiency~\cite{Dayarathna2024Deep}. Early approaches~\cite{Toda2022Lung} predominantly leveraged conditional generative adversarial networks (GANs), such as pix2pix~\cite{Aljohani2022Generating}, for paired image-to-image translation. Later on, CycleGAN, an unsupervised GAN framework that eliminates the requirement for paired training datasets, gained rapid popularity, particularly for translation between MRI and CT images~\cite{Mcnaughton2023Synthetic,Sun2023Double}. More recently, emerging architectures based on transformer and diffusion models have been increasingly adopted for medical modality translation tasks~\cite{Ji2024Diffusion,Phan2024Structural}, achieving higher structural fidelity and improved anatomical consistency.

Expanding beyond conventional medical image modalities, recent works have explored synthesizing X-ray images directly from natural optical images. Despite the substantial domain gap, such methods possess important clinical potential, notably reducing radiation exposure and facilitating rapid disease screening. For instance, \etal{Haiderbhai}~\cite{Haiderbhai2020Pix2xray} developed the ``pix2xray'' model to generate realistic hand X-rays from regular photographic images. Similarly, GAN-based frameworks have been adopted for synthesizing spinal X-ray images based on surface data~\cite{Teixeira2018Generating}, RGBD~\cite{Meng2023Radiograph} and RGB images~\cite{He2024Conditional}. While these approaches demonstrate the feasibility of translating optical imagery into medical imaging domains, they are primarily limited to image-domain transformations from single-view perspectives.


\subsection{Novel View Synthesis in Medical Imaging}
Synthesizing novel views of anatomical structures from limited medical imaging data is challenging due to the inherent depth ambiguity in 2D projections and the scarcity of multi-view medical datasets~\cite{Shen2022Novel}.
Despite these challenges, recent advances in deep learning have introduced strategies that integrate implicit or explicit 3D anatomical knowledge into image generation to address this issue.

One approach involves reconstructing approximate 3D representations from available data to enable the generation of images from arbitrary viewpoints. For instance, X2CT-GAN~\cite{Ying2019X2ct} reconstructs 3D chest CT volumes from two orthogonal X-ray images. This method employs a specially designed generator network to increase data dimensionality from 2D to 3D, facilitating the synthesis of novel views through projection of the reconstructed volume. Another strategy employs latent-space disentanglement of anatomical structure and viewpoint, allowing the synthesis of different perspectives by modifying viewpoint vectors. XraySyn~\cite{Peng2021Xraysyn} exemplifies this approach by embedding X-ray images into a latent space trained on CT data, allowing for rotation and forward projection to synthesize images from new perspectives.
Recent advancements have also seen the application of implicit 3D representations, such as Neural Radiance Fields (NeRF)~\cite{Corona2022Mednerf,Cai2024Structure}, and 3D Gaussian Splatting (3DGS)~\cite{Cai2024Radiative}, in medical imaging. These methods have shown promise in representing 3D structures from limited 2D projections. However, all these methods have been primarily confined to single-modality scenarios, and they have not yet been extended to cross-modality transformations.

\section{Method}
\label{sec:method}
In this section, we first present the study motivation and overview of the proposed LatXGen framework (Section~\ref{sec:method::subsec:Overview}). Next, we delve into the details of the major components of LatXGen, i.e., a data transformation module (Section~\ref{sec:method::subsec:data_trans}), and a dual-stage model consists of spine morphology estimation (SME) stage (Section~\ref{sec:method::subsec:spine_mph_est}) and lateral radiograph synthesis (LRS) stage (Section~\ref{sec:method::subsec:latgen}).

\begin{figure*}[!t]
    \centering
    \includegraphics[width=1.\linewidth]{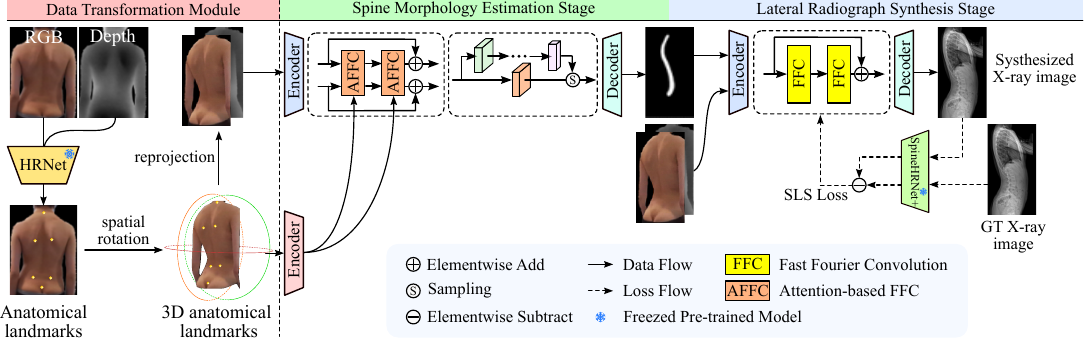}
    \caption{Framework overview of the proposed LatXGen which consists of a data preprocssing module and a dual-stage (spine morphology estimation stage and lateral radiograph synthesis stage) model.}
    \label{fig:framework}
    \vspace{-1em}
\end{figure*}

\subsection{{Overview}}
\label{sec:method::subsec:Overview}

\subsubsection{Motivation}
\label{sec:method::subsec:Overview::subsubsec:motivation}
Generating lateral X-ray images based on RGBD human back images need to resolve two crucial problems. 1) learning a precise cross-modality transformation between optical surface images and radiographic projections, and 2) accurately modeling anatomical variations of spinal structures across different viewpoints. Given that the accuracy of spinal morphology estimation under novel viewpoints directly influences the quality of lateral X-ray synthesis, we propose a dual-stage framework that progressively estimates lateral spinal morphology and synthesizes anatomically realistic lateral-view spinal X-ray images.

\subsubsection{Framework}
\label{sec:method::subsec:Overview::subsubsec:framework}
An overview of LatXGen framework is shown in Fig.~\ref{fig:framework}. Given posterior-view RGB and depth images $I_{RGB} \in \mathbb{R}^{H\times W\times 3}$ and $I_{D} \in \mathbb{R}^{H\times W}$, the objective is to synthesize the lateral spinal X-ray image $I_{X} \in \mathbb{R}^{H\times W}$, where $H$ and $W$ represent image height and width, respectively. Initially, 3D anatomical landmarks $v$ are extracted from the RGBD images using our self-developed model~\cite{Meng2023Radiograph}. These, along with the images, are rotated by an angle $\theta$ via the data transformation operation $M_T(\cdot)$, producing transformed RGBD and landmark data:
\begin{equation}
\label{equ:data_trans}
\left\{ I_{RGB}^\theta, I_{D}^\theta, v^\theta \right\} = M_T\left( \left\{ I_{RGB}, I_{D}, v \right\} \right).
\end{equation}

The transformed images and 3D landmarks are then fed into the SME stage $M_{SME}(\cdot)$ to generate the lateral spine curve map $\hat{I_{S}}$,
\begin{equation}
\hat{I_{S}} = M_{SME}\left( \left\{ I_{RGB}^{\theta}, I_{D}^{\theta}, v^{\theta} \right\}; \Theta_{SME} \right),
\end{equation}
where $\Theta_{SME}$ denotes the parameter set of this module. Here, $\hat{I}_S$ represents the model estimated spine curve map, while the ground truth $I_S$ is directly derived from segmentation of the original lateral radiograph, and neither $\hat{I}_S$ nor $I_S$ undergoes rotation. $M_{SME}(\cdot)$ mainly consists of a series of stacked Attention-based Residual FFC (ARFFC) blocks $\left\{ m_\alpha^t(\cdot) \right\}_{t=1,\cdots,T}$ and a Spatial Deformation Network (SDN) $m_\beta(\cdot)$. 
The input transformed images are first processed by a two-convolutional-layer encoder $\mathcal{E}_I$ to obtain the initial feature map $F_0=\left\{F_0^{g}, F_0^{l}\right\}$, which comprises of a global feature map $F_0^{g}$ and a local feature map $F_0^l$.
Simultaneously, the input 3D landmarks are processed by an encoder $\mathcal{E}_v$ comprising two concatenated fully-connected (FC) layers to obtain the landmark features $F^v=\mathcal{E}_v(v^\theta)$. The $t^\mathrm{th}$ ARFFC block takes the feature map $F_{t-1}$ as input and outputs the feature map $F_t$ $(t = 1,2,\cdots,T)$, which can be expressed as,
\begin{equation}
\label{equ:lrffc}
\begin{aligned}
F_t = \left\{ F_t^{g}, F_t^{l} \right\} = m_{\alpha}^t \left( \left\{ F_{t-1}^{g}, F_{t-1}^{l} \right\}, F^v; \Theta_\alpha^t \right),
\end{aligned}
\end{equation}
where $\Theta_\alpha^t$ is the parameters of the $t^\mathrm{th}$ ARFFC block. The output feature $F_T$ of the SME module is further processed by the SDN $m_\beta(\cdot)$ and a two-convolutional-layer decoder $\mathcal{D}$ to generate the lateral spine curve map $\hat{I_{S}}$,
\begin{equation}
\hat{I_{S}} = \mathcal{D} \circ m_\beta\left( F_T; \Theta_\beta \right),
\end{equation}
where $\Theta_\beta$ denotes the parameter of the SDN.

Finally, the estimated spine curve map $\hat{I_{S}}$, along with transformed images $I_{RGB}^\theta$ and $I_{D}^\theta$ are combined as inputs to the next LRS stage, i.e., $M_{LRS}(\cdot)$. The overall process can be represented as,
\begin{equation}
\hat{I_X} = M_{LRS}\left( \left\{I_{RGB}^\theta, I_{D}^\theta, \hat{I_{S}}\right\}; \Theta_{LRS} \right),
\end{equation}
where $\Theta_{LRS}$ denotes the parameters of the LRS module, and $\hat{I_X}$ is the synthesized lateral spinal X-ray image.

\subsection{Data Transformation}
\label{sec:method::subsec:data_trans}
Inspired by~\cite{Teixeira2018Generating}, anatomical landmarks offer valuable geometric priors for guiding image synthesis. To incorporate such structural cues, we introduce back surface landmark information into the generation process to enhance the estimation of lateral spinal morphology. As illustrated in Fig.~\ref{fig:framework}, we employ a previously developed model~\cite{Meng2023Radiograph} to predict 2D anatomical landmarks directly from RGBD images. These landmarks are then projected into 3D space using geometric depth information, yielding 3D anatomical keypoints on the back surface. Given the substantial viewpoint disparity between posterior and lateral images, directly estimating lateral spinal morphology from a posterior view is inherently challenging. To address this, we introduce a viewpoint transformation strategy that generates an intermediate view, providing enriched structural cues closer to the lateral perspective. Specifically, spatial rotation is performed on both the captured RGBD images $I_{RGB}$ and $I_{D}$, and the associated 3D anatomical landmarks $\mathbf{v}$, rotating them by a specified angle $\theta$. The rotated 3D data is then re-projected onto the camera imaging plane, generating novel-view images $I^\theta_{RGB}$ and $I^\theta_{D}$, and transformed landmarks $\mathbf{v}^\theta$. The resulting images $I^\theta_{RGB}$, $I^\theta_{D}$ and landmarks $\mathbf{v}^\theta$ are then used as inputs for the subsequent spine morphology estimation stage.

\begin{figure}[!t]
    \centering
    \includegraphics[width=1.\linewidth]{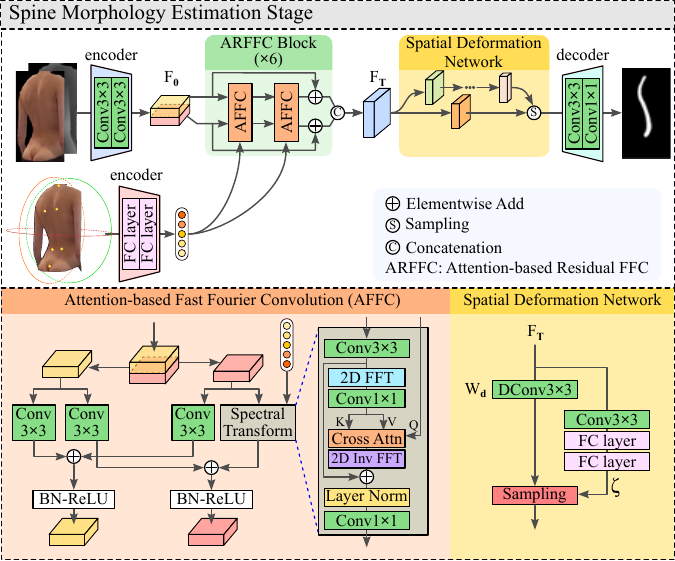}
    \caption{Detailed architecture of spine morphology estimation stage.}
    \label{fig:sme}
    \vspace{-2em}
\end{figure}

\subsection{Spine Morphology Estimation}
\label{sec:method::subsec:spine_mph_est}
This stage aims to accurately estimate lateral spine morphology by predicting a lateral spine curve map. As shown in Fig.~\ref{fig:sme}, the architecture mainly comprises six stacked ARFFC blocks, followed by an SDN that further refines the extracted features to generate the lateral spine curve map.

Each ARFFC block contains two concatenated AFFC modules. Specifically, within a AFFC module of the $t^\mathrm{th}$ ARFFC block, the input feature map $F_{t-1}$ is initially decomposed into local ($F_{t-1}^l$) and global ($F_{t-1}^g$) feature components, each undergoing separate, dedicated processing branches. Given that the local features do not require capturing long-range dependencies, the \textit{local branch} refines these local features by concurrently integrating both local and global features obtained from the previous $(t-1)^\mathrm{th}$ block using two parallel convolutional layers.
\begin{equation}
\label{equ:SME_local}
F_t^l=\sigma \left( \mathrm{Conv}\left(F_{t-1}^l\right) + \mathrm{Conv}\left(F_{t-1}^g\right) \right),
\end{equation}
where $\sigma$ denotes the batch normalization and ReLU operations and $\mathrm{Conv}(\cdot)$ stands for convolution operation. The \textit{global branch} refines the features obtained from the previous $(t-1)^\mathrm{th}$ block in the frequency domain and the resulting outputs combine with the refined local features processed by a convolutional layer to generate the new global features $F_{t}^g$.
\begin{equation}
\label{equ:SME_global}
F_t^g=\sigma \left( \mathrm{Conv}\left(F_{t-1}^l\right) + \eta\left(F_{t-1}^g\right) \right),
\end{equation}
where $\eta(\cdot)$ stands for the composite operations within the \textit{spectral transform} module. In contrast to existing spectral transform architectures~\cite{Suvorov2022Resolution}, we incorporate 3D anatomical landmark knowledge into the global frequency domain features via a cross-attention mechanism. Specifically, the input global features $F_{t-1}^g$ are first transformed into the frequency domain to expand the receptive field~\cite{Sinha2022NL}, and subsequently processed by a convolutional layer and further splitted into key $K_t$ and value $V_t$ maps, while the landmark features $F^v$ are involved in the attention mechanism as the query map across all blocks. As a result, the process of cross-attention calculation can be formulated as,
\begin{align}
F_f^g &= \mathrm{Softmax}(\frac{K_t \otimes Q_t^\mathrm{T}}{\sqrt{d}}) \otimes V_t
\end{align}
where $\otimes$ means matrix multiplication, $\circ$ denotes function composition and $\frac{1}{\sqrt{d}}$ serves as a scaling factor that regulates the magnitude of the dot product between $K_t$ and $Q_t$, and $F_f^g$ denotes the frequency global feature map. Subsequently, $F_f^g$ is transformed back to the spatial domain to produce the outputs of the spectral transform module.
\begin{equation}
\eta(F_{t-1}^g) = \mathrm{Conv}\left(\mathcal{F}^{-1}\left( F_f^g\right) + \mathrm{Conv}\left(F_{t-1}^g\right) \right)
\end{equation}

The final output $F_T$ from the last ARFFC block integrates both global and local image features, along with anatomical landmark information. Since the input images are not acquired from true lateral views, the extracted features require further spatial refinement. To this end, we introduce the SDN model that refines features by modeling both deformation and spatial translation. Inspired by the architecture in~\cite{Jaderberg2015Spatial}, our SDN incorporates deformable convolutions to enhance the network’s capacity for learning non-rigid transformations. As illustrated in Fig.~\ref{fig:sme}, $F_T$ is processed through a deformable convolution layer to extract deformation-aware features. In parallel, two FC layers learn sampling parameters, which are then used to sample and refine the features, improving the robustness of the model to spatial misalignments. Therefore, the operations within SDN can be formulated as,
\begin{equation}
\hat{I_{S}} = \mathcal{D} \circ \mathrm{Sample}\left(F_T * W_d, \zeta \right),
\end{equation}
where $\zeta = \mathrm{Linear}\circ\mathrm{Conv}(F_T)$. $*$ denotes convolution, $W_d$ is the weight of the deformable convolution layer, and $\mathrm{Linear}(\cdot)$ represents the linear mapping performed by the two FC layers.

\begin{figure}[!t]
    \centering
    \includegraphics[width=1.\linewidth]{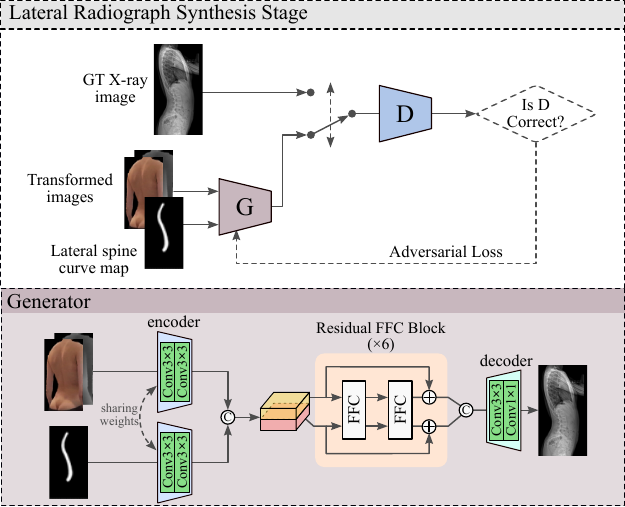}
    \caption{Detailed framework of lateral radiograph synthesis stage.}
    \label{fig:lrs}
    \vspace{-2em}
\end{figure}

\subsection{Lateral Radiograph Synthesis}
\label{sec:method::subsec:latgen}
The objective of LRS stage is to synthesize realistic lateral spinal X-ray images based on the transformed images and the predicted lateral spine curve map obtained from the preceding SME stage.

As illustrated in Fig.~\ref{fig:lrs}, the LRS stage employs a generative adversarial network (GAN)-based architecture. Many of its foundational modules, such as the encoder and decoder, share similar structures with those used in the preceding SME stage.
Following initial feature extraction by the encoder, the generator employs six FFC residual blocks, each comprising two FFC layers with a configuration consistent with~\cite{Suvorov2022Resolution}. The final FFC residual block produces high-level features, which are then passed through the decoder to generate the synthesized lateral spinal radiographs. The entire process can be formulated as,
\begin{equation}
\hat{I}_X = \mathcal{D}\circ \varphi \circ\mathcal{E}_I\left(\left\{I_{RGB}^\theta, I_D^\theta, \hat{I}_S\right\}\right),
\end{equation}
where $\mathcal{E}_I(\cdot)$ and $\mathcal{D}(\cdot)$ represent the encoder and decoder, respectively, and $\varphi(\cdot)$ denotes the mapping performed by the six stacked FFC residual blocks.

\section{Experiments}
\subsection{Experimental Setups}
\label{sec:exp::subsec:expsetups}

\subsubsection{Datasets}
This study used proprietary dataset collected from Queen Mary Hospital (QMH) and Duchess of Kent Children's Hospital (DKCH) in Hong Kong from October 2019 to August 2023. All enrolled participants were adolescents (aged 10–18) clinically diagnosed with AIS. Exclusion criteria included psychological disorders, history of trauma affecting posture or mobility, severe dermatological conditions interfering with optical imaging, and any known oncological conditions. Ethical approval for this study was granted by the Institutional Review Board of The University of Hong Kong (protocol No.UW22-270 and No.UW19-620), and informed consent was provided by all participants or their guardians prior to data collection.

Each participant underwent standardized imaging procedures, including acquisition of a posterior unclothed RGBD back surface image and a corresponding standing lateral full-spine radiograph. A total of 3,264 paired RGBD back surface images and lateral spinal radiographs were collected, comprising predominantly female subjects (72.97\%). The dataset was randomly partitioned into a training set of 2,611 image pairs and an independent testing set containing 653 pairs.

\subsubsection{Data Labelling}
Senior spinal surgeons manually annotated critical anatomical landmarks, specifically identifying the spinous process of the seventh cervical vertebra (C7) and the tip of the coccyx (ToC) on both RGBD images and lateral spinal X-rays.These landmarks served as reference points to define the spinal range and align the RGBD back surface images with the corresponding lateral radiographs. SpineHRNet+~\cite{Meng2022Artificial} was used to produce the spine curve map for each X-ray image.
Additionally, three sagittal spinal alignment parameters were manually measured on lateral spinal X-rays: namely thoracic kyphosis angle (TKA), lumbar lordosis angle (LLA), and sacral slope angle (SSA), on radiographs. Specifically, TKA was measured between the superior endplate of the fifth thoracic vertebra (T5) and the inferior endplate of the twelfth thoracic vertebra (T12); LLA was measured between the superior endplates of the first lumbar vertebra (L1) and first sacral vertebra (S1); SSA was defined by the angle formed between the horizontal plane and the superior endplate of S1. Clinically accepted normal ranges for these parameters are TKA: $20^\circ–40^\circ$, LLA: $20^\circ–45^\circ$, and SSA: $32^\circ–49^\circ$~\cite{Meng2022Artificial}.

\subsubsection{Training Details}
The input RGB and depth images were cropped and resized to $300\times400$ patches to cover the entire spine region. Data augmentation included horizontal flipping, shifting, and random rotations within $[0^\circ, 5^\circ]$. The two stages of LatXGen were trained separately. Both were optimized using the ADAM optimizer (default settings) for 400 epochs with a batch size of 10. The initial learning rate was set to $10^{-3}$ and gradually decreased to $10^{-5}$ via the cosine decay strategy. All experiments were conducted on a system equipped with an Intel Xeon Platinum 8373C CPU (2.60GHz) and an NVIDIA GeForce RTX 3090 (24GB) GPU.

\subsection{Training Loss}
\label{sec:experiments::subsec:loss}
\subsubsection{Adversarial Loss}
To ensure high-quality image generation, both the SME and LRS stages adopt adversarial learning. We define discriminators $D_{S}$ and $D_{L}$ for the SME and LRS stages, respectively. These discriminators share the same structure as PatchGAN~\cite{Isola2017Image}, to differentiate between real and synthesized data. The corresponding generators are denoted as $G_{S}$ and $G_{L}$. For SME stage, the adversarial losses are:

{\small
\vspace{-1em}
\begin{align}
\mathcal{L}_{D_{S}} &= \mathbb{E}_{I_{S}}\left[\log D_{S}\left(I_{S}\right)\right] + \mathbb{E}_{I^{\theta}}\left[\log(1 - D_{S}(G_{S}(I^{\theta}))) \right], \\
\mathcal{L}_{G_{S}} &= \mathbb{E}_{I^{\theta}}[\log(1 - D_{S}(G_{S}(I^{\theta})))],
\end{align}}

\noindent where $I_S$ is the spine curve ground truth (GT), $\mathbb{E}[\cdot]$ denotes expectation operation, and for simplicity we use $I^\theta$ to represents $\left\{I_{RGB}^\theta, I_{D}^\theta, I_{v}^\theta \right\}$. Similarly, for the LRS stage, the adversarial losses are defined as:

{\small
\vspace{-1em}
\begin{align}
\mathcal{L}_{D_{L}} &= \mathbb{E}_{I_{X}}\left[\log D_{L}\left(I_{X}\right)\right] + \mathbb{E}_{I_{m}}\left[\log(1 - D_{S}(G_{S}(I_{m}))) \right] \\
\mathcal{L}_{G_{L}} &= \mathbb{E}_{I{in}}[\log(1 - D_{S}(G_{S}(I_{m})))],
\end{align}}

\noindent where $I_m$ denotes the inputs of LRS stage $\left\{ I_{RGB}^\theta, I_D^\theta,\hat{I_S} \right\}$.
\subsubsection{L1 Loss}
To improve pixel-level accuracy, L1 loss is employed in both modules:
\begin{align}
\mathcal{L}_1^{SME} &= \frac{1}{N}\sum_{n=1}^N \left| \hat{I_S} - I_S \right| \\
\mathcal{L}_1^{LRS} &= \frac{1}{N}\sum_{n=1}^N \left| \hat{I_X} - I_X \right|,
\end{align}
where $I_X$ denotes the real lateral spinal X-ray image. $N$ represents the batchsize.
\subsubsection{Spine Landmark Supervision Loss}
To enhance anatomical accuracy, we define a novel spine landmark supervision (SLS) loss in the final LRS stage. A pre-trained SpineHRNet+~\cite{Meng2022Artificial} model extracts landmark features from both predicted and GT X-rays as inputs. We compute the feature difference using intermediate feature maps obtained from the layer preceding the model's head layer to acquire the SLS loss,
\begin{equation}
\mathcal{L}_{SLS} = \frac{1}{N}\sum_{n=1}^N\left(f(\hat{I}_X) - f(I_X)\right)^2,
\end{equation}
where $f(\cdot)$ denotes the feature extractor part of SpineHRNet+. Accordingly, the final loss functions for the SME and LRS stages are defined as:
\begin{align}
\mathcal{L}_{SME} &= \mathcal{L}_{G_S} + \alpha\mathcal{L}_1^{SME}, \\
\mathcal{L}_{LRS} &= \mathcal{L}_{G_L} + \beta\mathcal{L}_1^{LRS} + \gamma\mathcal{L}_{SLS},
\end{align}
where $\alpha$, $\beta$ and $\gamma$ were empirically set to $0.5$, $0.5$ and $3$, respectively.

\section{Results and Discussion}
\label{sec:result_disc}
This section presents a comprehensive evaluation of the proposed LatXGen framework across three key tasks: (1) spine curve generation, (2) lateral spinal radiograph synthesis, and (3) sagittal alignment parameter prediction. Given the limited existing work on cross-modality novel view generation, direct comparisons are challenging. As both stages of LatXGen are GAN-based, we primarily benchmark our method against advanced GAN-based models to validate the effectiveness of the overall pipeline and key submodules in each stage.
Additionally, a series of ablation studies are conducted to assess the impact of optical rotation angle and the contribution of the proposed SLS loss.

\begin{figure}[!t]
    \centering
    \includegraphics[width=.8\linewidth]{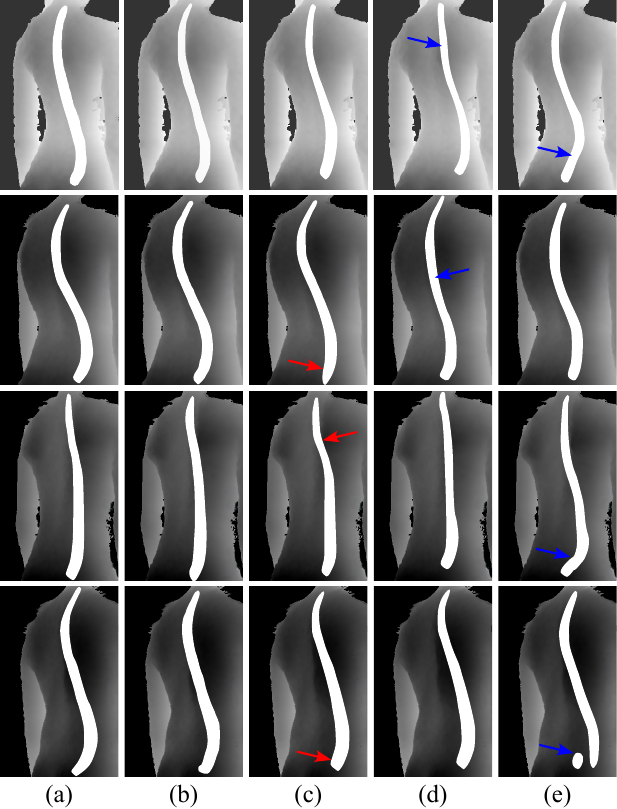}
    \caption{Visual comparison of spine curve map generation. The white mask denote the spine region. The five columns are: (a) spine curve GT, (b) SME stage with SDN, (c) SME stage without SDN, (d) Pix2PixGAN, (e) Pix2PixHDGAN, respectively.}
    \label{fig:spinecurvegen}
    \vspace{-1.em}
\end{figure}
\begin{table}[!t]
\caption{Quantitative comparison of our LatXGen against advanced GAN-based frameworks on spine curve generation.}
\label{table:spinecurvegen}
\addtolength{\tabcolsep}{-3pt}
\begin{tabular}{l|ccccc}
\hline
\textbf{Model} & \textbf{Accuracy} & \textbf{Precision} & \textbf{Sensitivity} & \textbf{F1-score} & \textbf{IoU}   \\ \hline\hline
Pix2PixGAN~\cite{Isola2017Image}     & 0.973             & 0.653              & 0.642                & 0.646             & 0.495          \\
Pix2PixHDGAN~\cite{Wang2018High}   & 0.978             & 0.690              & 0.663                & 0.674             & 0.525          \\
SME w/o SDN    & \textbf{0.982}    & 0.750              & 0.710                & 0.727             & 0.586          \\
SME with SDN   & \textbf{0.982}    & \textbf{0.757}     & \textbf{0.712}       & \textbf{0.732}    & \textbf{0.592} \\
\hline
\end{tabular}
\vspace{-1em}
\end{table}

\subsection{Spine Curve Generation}
\label{sec:result_disc::subsec:SCG}

We evaluate the performance of our proposed SME module against two advanced GAN-based models, namely, Pix2PixGAN~\cite{Isola2017Image} and Pix2PixHDGAN~\cite{Wang2018High}.To ensure fair comparison, all models were trained under identical conditions, including consistent input/output formats, discriminator architectures, and loss items. Table~\ref{table:spinecurvegen} reports the quantitative results for lateral spine curve generation. Our SME module, particularly when integrated with the SDN, achieves the best overall performance.

Notably, even without SDN, our generator based on the proposed ARFFC blocks significantly outperforms the traditional ResNet-based architectures used in Pix2PixGAN and Pix2PixHDGAN, highlighting the effectiveness of the ARFFC design. Fig.~\ref{fig:spinecurvegen} presents the visual comparison of generated spine curves. The ARFFC-based generator better preserves the global curvature trend, as illustrated by the blue arrows (\blue{$\mathbf{\rightarrow}$}) in Fig.~\ref{fig:spinecurvegen}, which highlight areas where compared models deviate from the ground truth. Furthermore, the red arrows (\red{$\mathbf{\rightarrow}$}) in Fig.~\ref{fig:spinecurvegen} mark areas with insufficient detail reconstruction. The inclusion of SDN further enhances local curve accuracy, demonstrating its utility in modeling fine-grained geometric deformations. The SME module demonstrates high efficiency, requiring on average 0.05 seconds per image and approximately 2.5 GB of GPU memory. Despite SME ensuring alignment with RGBD inputs, curve estimation errors (under- or overestimation) may still propagate into the synthesized radiographs and affect anatomical fidelity.

\begin{figure}[!t]
    \centering
    \includegraphics[width=1.\linewidth]{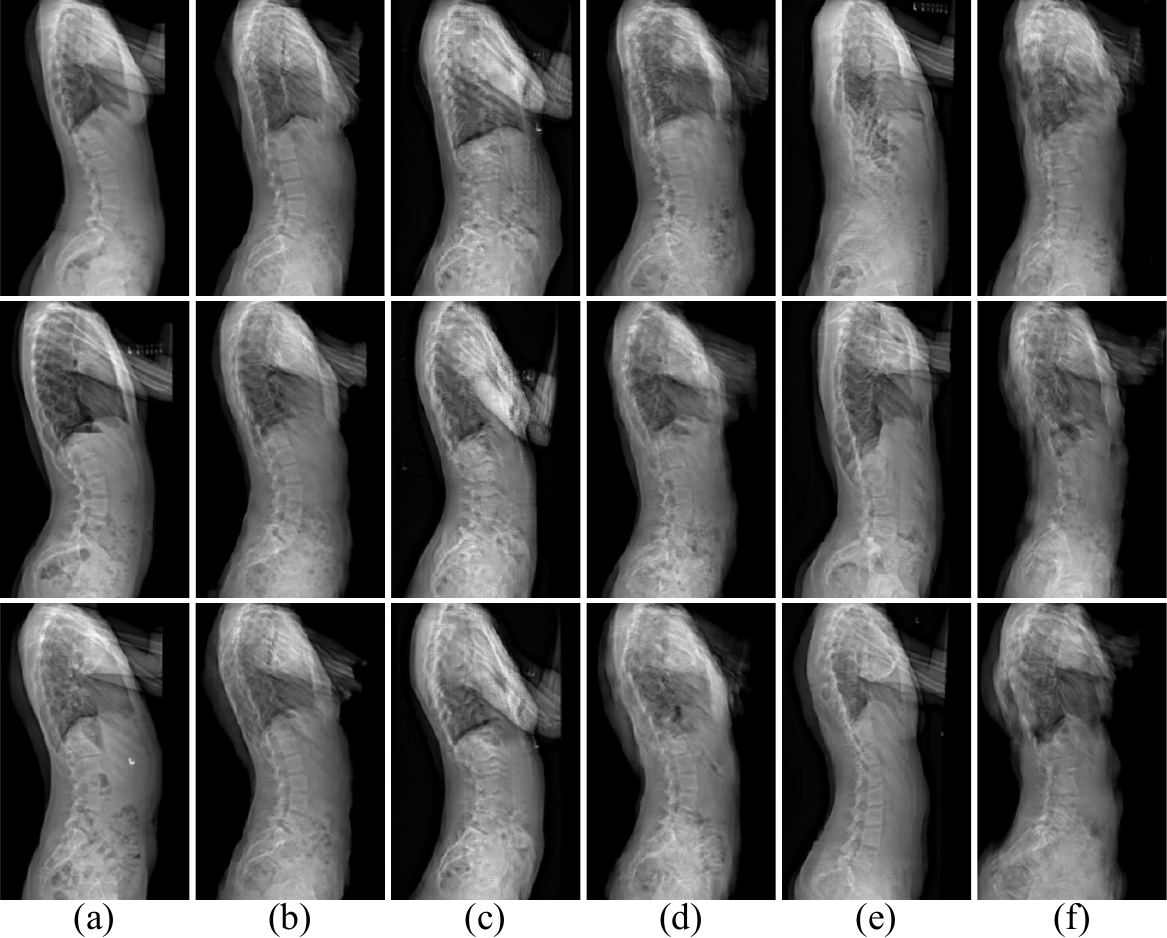}
    \caption{Visual comparison of lateral spinal radiograph synthesis. The six columns are: (a) GT radiograph, (b) LatXGen, (c) LatXGen replace the LRS with Pix2PixGAN, (d) LatXGen replace the LRS with Pix2PixHDGAN, (e) Pix2PixGAN, and (f) Pix2PixHDGAN.}
    \label{fig:xraygen}
    \vspace{-.5em}
\end{figure}

\begin{figure}[!t]
    \centering
    \includegraphics[width=1.\linewidth]{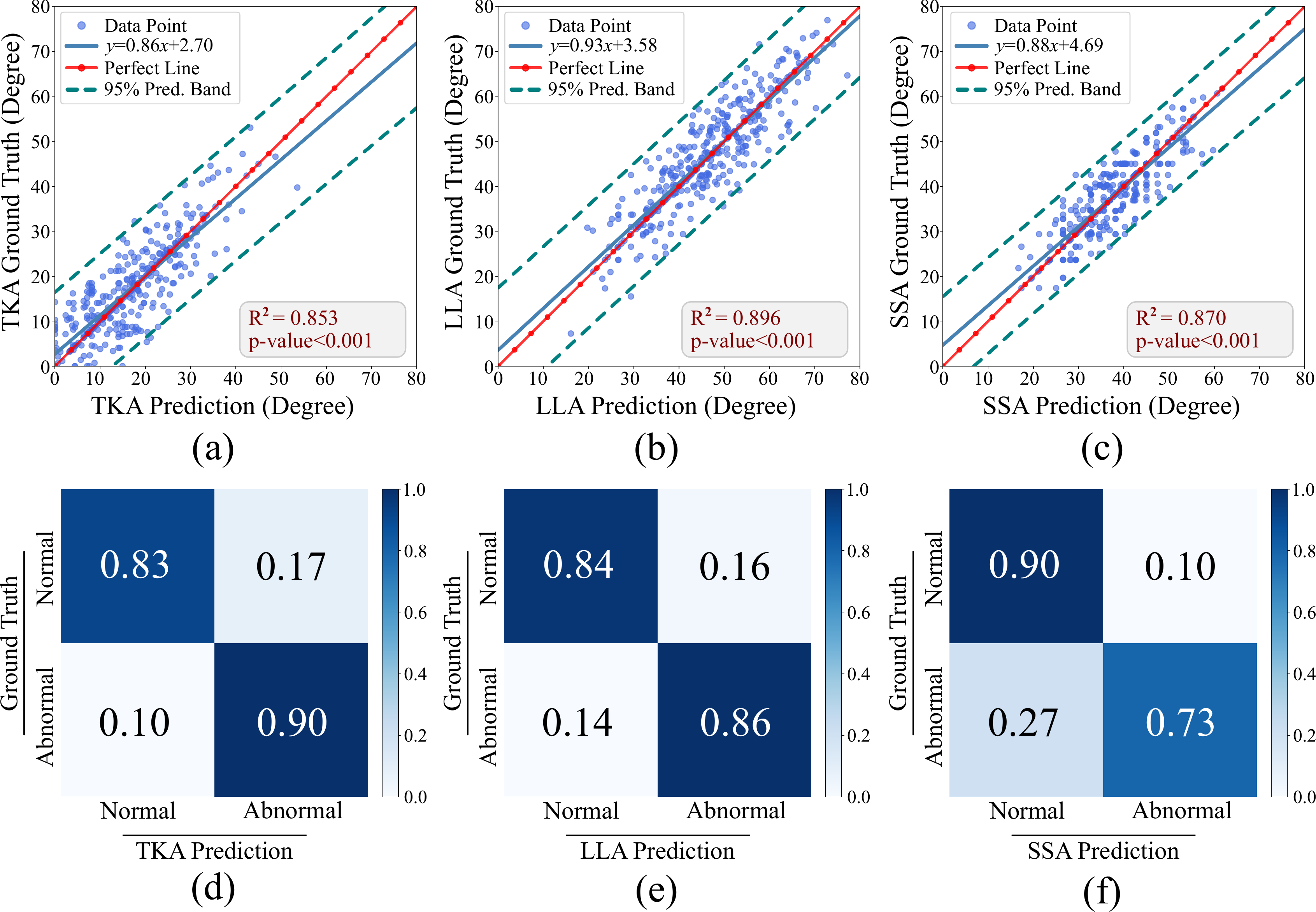}
    \caption{Linear regression and confusion matrix results on sagittal spine alignment parameter prediction. (a)-(c) present the linear regression results on TKA, LLA and SSA prediction, respectively. (d)-(f) show the confusion matrix results on sagittal curve detection.}
    \label{fig:lr_cm}
    \vspace{-1.5em}
\end{figure}

\subsection{Lateral Spinal Radiograph Synthesis}
Fig.~\ref{fig:xraygen} presents a visual comparison of lateral spinal radiographs synthesized by different generator architectures. To evaluate the effectiveness of the proposed LRS module, we retained the overall LatXGen framework and varied only the generator within the LRS stage, testing three configurations: our proposed LRS module, Pix2PixGAN, and Pix2PixHDGAN. Their respective results are shown in Fig.~\ref{fig:xraygen} columns (b), (c), and (d). As shown, the proposed generator in LRS stage produces more anatomically realistic results in the spinal regions, with clearer vertebral boundaries.
In contrast, Fig.~\ref{fig:xraygen} columns (e) and (f) display results from standalone Pix2PixGAN and Pix2PixHDGAN models applied directly to lateral spinal radiograph synthesis. These outputs exhibit noticeable structural distortions, including exaggerated vertebral wedging and, in the case of Pix2PixHDGAN, failure to capture coherent spinal anatomy (the generated spinal regions are blurred).
Comparing columns (c) vs. (e) and (d) vs. (f) highlights the advantage and benefits of LatXGen's progressive architecture, i.e., first estimating the spine curve and then synthesizing the radiograph based on the curve. The dual-stage generation strategy enables more accurate modeling of sagittal spinal morphology. Quantitative metrics evaluating the image generation quality across the different models and configurations are presented in Table~\ref{table:xraygen}. The LRS module processes each image in about 0.05 seconds on average and consumes about 0.8 GB of GPU memory. Although these measures are widely used benchmarks, they may be affected by non-spinal regions. To highlight clinical relevance, we also report sagittal parameters (TKA, LLA, SSA) from the synthesized radiographs as complementary validation of the model’s effectiveness.

\begin{table}[!t]
\caption{Quantitative evaluation of image quality of the synthesized radiographs in terms of different models and configurations presented in Fig.~\ref{fig:xraygen}}.
\label{table:xraygen}
\addtolength{\tabcolsep}{-1pt}
\begin{tabular}{c|cccc|ccc}
\hline
\multirow{2}{*}{\textbf{Pipeline}} & \multicolumn{4}{c|}{\textbf{Model}}               & \multicolumn{3}{c}{\textbf{Metrics}}                                      \\ \cline{2-8} 
    & \multicolumn{1}{l}{SME}           &   LSR                              &   \cite{Isola2017Image}                  &   \cite{Wang2018High}                & \textbf{LPIPS}$\downarrow$ & \textbf{FID}$\downarrow$                & \textbf{PSNR}$\uparrow$              \\ \hline\hline
(b) & \darkspringgreen{$\newcheckmark$} & \darkspringgreen{$\newcheckmark$} &                                    &                                   & \textbf{0.133} & \textbf{93.970}             & \textbf{20.854}            \\
(c) & \darkspringgreen{$\newcheckmark$} &                                   & \darkspringgreen{$\newcheckmark$}  &                                   & 0.197          & 98.711                      & \multicolumn{1}{c}{18.837} \\
(d) & \darkspringgreen{$\newcheckmark$} &                                   &                                    & \darkspringgreen{$\newcheckmark$} & 0.188          & 96.970                      & 19.071                     \\
(e) &                                   &          &  \darkspringgreen{$\newcheckmark$} &         & 0.205          & 140.895                     & 18.870                     \\
(f) &                                   &          &         & \darkspringgreen{$\newcheckmark$} & 0.272          & \multicolumn{1}{l}{149.138} & 16.153                     \\ \hline
\end{tabular}
\end{table}

\begin{table}[!t]
\caption{Quantitative results on spine curve generation in terms of different rotation angles.}
\label{table:rotation_ablation}
\centering
\begin{tabular}{c|ccccc}
\hline
\textbf{Angle} & \textbf{Accuracy}       & \textbf{Precision}      & \textbf{Sensitivity}    & \textbf{F1 score}       & \textbf{IoU}            \\ \hline\hline
$0^\circ$  & 0.965            & \textbf{0.790}          & 0.550        0.649         &0.492        \\ 
$30^\circ$  & 0.970           & 0.613          & 0.698          & 0.631          & 0.497          \\ 
$45^\circ$   & \textbf{0.982} & 0.757 & 0.712          & \textbf{0.732} & \textbf{0.592} \\ 
$60^\circ$   & 0.975          & 0.609          & \textbf{0.714} & 0.655          & 0.502          \\ \hline
\end{tabular}
\vspace{-1.em}
\end{table}

\begin{table}[!t]
\caption{Quantitative evaluation on the effect of SLS loss on lateral spinal radiograph synthesis.}
\label{table:sls_ablation}
\centering
\begin{tabular}{l|lll}
\hline
\textbf{Loss}  & \textbf{LPIPS}$\downarrow$  & \textbf{FID}$\downarrow$  & \textbf{PSNR}$\uparrow$    \\ \hline\hline
with SLS       & \textbf{0.133} & \textbf{93.970} & \textbf{20.854} \\ 
w/o SLS        & 0.150          & 95.768          & 19.315          \\ \hline
\end{tabular}
\vspace{-1.5em}
\end{table}

\subsection{Sagittal Alignment Parameter Prediction}
\label{sec:results::subsec:sagittalparamprediction}
To assess the clinical applicability of the synthesized radiographs, we performed a comparative analysis between key sagittal alignment parameters (namely, TKA, LLA and SSA) measured on real radiographs and those derived directly from synthesized radiographs using SpineHRNet+~\cite{Meng2022Artificial}. Evaluation was conducted using linear regression and confusion matrix analysis, with results illustrated in Fig.~\ref{fig:lr_cm}. Fig.~\ref{fig:lr_cm}~\blue{(a)–(c)} plot predicted angles on synthesized radiographs (x-axis) against GT angles measured from real radiographs (y-axis). The regression results demonstrate strong alignment between predicted and GT values, with coefficient of determination ($R^2$) values of $0.853$ for TKA, $0.896$ for LLA, and $0.870$ for SSA. These high correlations confirm the effectiveness of the proposed method for quantitative assessment of sagittal spinal alignment. The mean prediction errors were –0.459$^\circ$ for TKA (95\% CI: –1.007$^\circ$ to 0.089$^\circ$), –0.300$^\circ$ for LLA (95\% CI: –0.823$^\circ$ to 0.223$^\circ$), and –0.107$^\circ$ for SSA (95\% CI: –0.517$^\circ$ to 0.302$^\circ$), confirming that the observed differences are small.

Fig.~\ref{fig:lr_cm}~\blue{(d)-(f)} present the confusion matrix results for classifying abnormal sagittal alignment parameters. Among the three metrics, SSA exhibits relatively lower specificity ($0.73$), which may be attributed to the greater presence of soft tissue in the sacral region, making it more challenging to extract accurate skeletal features from RGBD surface data. In contrast, both TKA and LLA demonstrate high sensitivity and specificity (TKA: sensitivity=$0.83$, specificity=$0.90$; LLA: sensitivity=$0.84$, specificity=$0.86$), indicating that the proposed method is highly effective in identifying sagittal spinal malalignment in the thoracic and lumbar regions.

\subsection{Ablation study}
\label{sec:results::subsec:ablation}
To determine the optimal rotation angle for the data transformation module, we evaluated spine curve generation performance under rotation angles of $0^\circ$, $30^\circ$, $45^\circ$, and $60^\circ$. As shown in Table~\ref{table:rotation_ablation}, a rotation of $45^\circ$ yielded the most accurate spine curve predictions. 

Table~\ref{table:sls_ablation} presents the effect of incorporating the SLS loss. Models trained with SLS achieved better perceptual quality (lower LPIPS: 0.133 vs. 0.150, lower FID: 93.970 vs. 95.768) and improved reconstruction accuracy (higher PSNR: 20.854 vs. 19.315). These results demonstrate that SLS effectively enhances the anatomical fidelity and visual realism of synthesized lateral spinal radiographs.

\section{Conclusion and Future Work}
In this study, we proposed LatXGen, a novel dual-stage generative framework for synthesizing anatomically realistic lateral spinal radiographs from posterior RGBD back images. The framework integrates attention-based FFC module (AFFC) and residual blocks (ARFFC) and an SDN model to integrate knowledge from both landmark and RGBD, and capture both global and local spinal morphology. Additionally, we introduced the SLS loss to enhance anatomical fidelity. Extensive experiments demonstrated that LatXGen outperforms existing advanced GAN-based models in spine curve estimation and radiograph synthesis. Its accurate performance in sagittal alignment parameter prediction underscores the potential of our approach for accurate, radiation-free assessment of sagittal spinal alignment in AIS. Nonetheless, our dataset is predominantly composed of female adolescents and was collected from two sites within the same geographic and institutional context. Future studies should validate LatXGen on broader populations, diverse clinical settings, and varied imaging protocols to further confirm its generalizability across different patient groups and spinal pathologies (e.g., Scheuermann’s disease).

\section*{Reference}
\bibliographystyle{IEEEtran}
\bibliography{IEEEabrv,reference}

\begin{thebibliography}{10}
\providecommand{\url}[1]{#1}
\csname url@samestyle\endcsname
\providecommand{\newblock}{\relax}
\providecommand{\bibinfo}[2]{#2}
\providecommand{\BIBentrySTDinterwordspacing}{\spaceskip=0pt\relax}
\providecommand{\BIBentryALTinterwordstretchfactor}{4}
\providecommand{\BIBentryALTinterwordspacing}{\spaceskip=\fontdimen2\font plus
\BIBentryALTinterwordstretchfactor\fontdimen3\font minus \fontdimen4\font\relax}
\providecommand{\BIBforeignlanguage}[2]{{%
\expandafter\ifx\csname l@#1\endcsname\relax
\typeout{** WARNING: IEEEtran.bst: No hyphenation pattern has been}%
\typeout{** loaded for the language `#1'. Using the pattern for}%
\typeout{** the default language instead.}%
\else
\language=\csname l@#1\endcsname
\fi
#2}}
\providecommand{\BIBdecl}{\relax}
\BIBdecl

\bibitem{Fong2015Population}
D.~Y. Fong \emph{et~al.}, ``A population-based cohort study of 394,401 children followed for 10 years exhibits sustained effectiveness of scoliosis screening,'' \emph{The Spine Journal}, vol.~15, no.~5, pp. 825--833, 2015.

\bibitem{Mak2021Patterns}
T.~Mak \emph{et~al.}, ``Patterns of coronal and sagittal deformities in adolescent idiopathic scoliosis,'' \emph{BMC Musculoskeletal Disorders}, vol.~22, 2021.

\bibitem{Cheung2018Curve}
J.~P.~Y. Cheung \emph{et~al.}, ``Curve progression in adolescent idiopathic scoliosis does not match skeletal growth,'' \emph{Clinical Orthopaedics and Related Research}, vol. 476, no.~2, pp. 429--436, 2018.

\bibitem{Weinstein2013Effects}
S.~L. Weinstein \emph{et~al.}, ``Effects of bracing in adolescents with idiopathic scoliosis,'' \emph{New England Journal of Medicine}, vol. 369, no.~16, pp. 1512--1521, 2013.

\bibitem{Chung2018Spinal}
N.~Chung \emph{et~al.}, ``Spinal phantom comparability study of cobb angle measurement of scoliosis using digital radiographic imaging,'' \emph{Journal of Orthopaedic Translation}, vol.~15, pp. 81--90, 2018.

\bibitem{Knott2014Sosort}
P.~Knott \emph{et~al.}, ``Sosort 2012 consensus paper: reducing x-ray exposure in pediatric patients with scoliosis,'' \emph{Scoliosis}, vol.~9, no.~1, p.~4, 2014.

\bibitem{Cote1998Study}
P.~C{\^o}t{\'e} \emph{et~al.}, ``A study of the diagnostic accuracy and reliability of the scoliometer and adam's forward bend test,'' \emph{Spine}, vol.~23, no.~7, pp. 796--802, 1998.

\bibitem{Choi2017CNN}
R.~CHOI \emph{et~al.}, ``Cnn-based spine and cobb angle estimator using moire images,'' \emph{IIEEJ Trans on Image Electronics and Visual Computing}, vol.~5, no.~2, pp. 135--144, 2017.

\bibitem{Zhang2023Deep}
T.~Zhang \emph{et~al.}, ``Deep learning model to classify and monitor idiopathic scoliosis in adolescents using a single smartphone photograph,'' \emph{JAMA Network Open}, vol.~6, no.~8, pp. e2\,330\,617--e2\,330\,617, 2023.

\bibitem{Roriz2021Automotive}
R.~Roriz, J.~Cabral, and T.~Gomes, ``Automotive lidar technology: A survey,'' \emph{IEEE TITS}, vol.~23, no.~7, pp. 6282--6297, 2021.

\bibitem{Potts2024Lidar}
M.~A. Potts, ``Lidar and x-ray: A retrospective comparison of spinal alignment,'' \emph{Medical Research Archives}, vol.~12, no.~9, 2024.

\bibitem{Kokabu2021Algorithm}
T.~Kokabu \emph{et~al.}, ``An algorithm for using deep learning convolutional neural networks with three dimensional depth sensor imaging in scoliosis detection,'' \emph{The Spine Journal}, vol.~21, no.~6, pp. 980--987, 2021.

\bibitem{Xu2020Back}
Z.~Xu \emph{et~al.}, ``Back shape measurement and three-dimensional reconstruction of spinal shape using one kinect sensor,'' in \emph{IEEE ISBI}, 2020, pp. 1--5.

\bibitem{Seoud2010Prediction}
L.~Seoud \emph{et~al.}, ``Prediction of scoliosis curve type based on the analysis of trunk surface topography,'' in \emph{IEEE ISBI}, 2010, pp. 408--411.

\bibitem{Meng2023Radiograph}
N.~Meng \emph{et~al.}, ``Radiograph-comparable image synthesis for spine alignment analysis using deep learning with prospective clinical validation,'' \emph{eClinicalMedicine}, vol.~61, 2023.

\bibitem{He2024Conditional}
Z.~He \emph{et~al.}, ``Conditional generative adversarial network-assisted system for radiation-free evaluation of scoliosis using a single smartphone photograph: a model development and validation study,'' \emph{eClinicalMedicine}, vol.~75, 2024.

\bibitem{Dayarathna2024Deep}
S.~Dayarathna, K.~T. Islam, S.~Uribe, G.~Yang, M.~Hayat, and Z.~Chen, ``Deep learning based synthesis of mri, ct and pet: Review and analysis,'' \emph{Med. Image Analy.}, vol.~92, p. 103046, 2024.

\bibitem{Toda2022Lung}
R.~Toda \emph{et~al.}, ``Lung cancer ct image generation from a free-form sketch using style-based pix2pix for data augmentation,'' \emph{Scientific Reports}, vol.~12, no.~1, p. 12867, 2022.

\bibitem{Aljohani2022Generating}
A.~Aljohani and N.~Alharbe, ``Generating synthetic images for healthcare with novel deep pix2pix gan,'' \emph{Electronics}, vol.~11, no.~21, p. 3470, 2022.

\bibitem{Mcnaughton2023Synthetic}
J.~McNaughton \emph{et~al.}, ``Synthetic mri generation from ct scans for stroke patients,'' \emph{BioMedInformatics}, vol.~3, no.~3, pp. 791--816, 2023.

\bibitem{Sun2023Double}
B.~Sun \emph{et~al.}, ``Double u-net cyclegan for 3d mr to ct image synthesis,'' \emph{International Journal of Computer Assisted Radiology and Surgery}, vol.~18, no.~1, pp. 149--156, 2023.

\bibitem{Ji2024Diffusion}
W.~Ji and A.~C. Chung, ``Diffusion-based domain adaptation for medical image segmentation using stochastic step alignment,'' in \emph{MICCAI}, 2024, pp. 188--198.

\bibitem{Phan2024Structural}
V.~M.~H. Phan, Y.~Xie, B.~Zhang, Y.~Qi, Z.~Liao, A.~Perperidis, S.~L. Phung, J.~W. Verjans, and M.-S. To, ``Structural attention: Rethinking transformer for unpaired medical image synthesis,'' in \emph{MICCAI}, 2024, pp. 690--700.

\bibitem{Haiderbhai2020Pix2xray}
M.~Haiderbhai, S.~Ledesma, S.~C. Lee, M.~Seibold, P.~F{\"u}rnstahl, N.~Navab, and P.~Fallavollita, ``pix2xray: converting rgb images into x-rays using generative adversarial networks,'' \emph{International Journal of Computer Assisted Radiology and Surgery}, vol.~15, pp. 973--980, 2020.

\bibitem{Teixeira2018Generating}
B.~Teixeira, V.~Singh, T.~Chen, K.~Ma, B.~Tamersoy, Y.~Wu, E.~Balashova, and D.~Comaniciu, ``Generating synthetic x-ray images of a person from the surface geometry,'' in \emph{IEEE CVPR}, 2018, pp. 9059--9067.

\bibitem{Shen2022Novel}
L.~Shen \emph{et~al.}, ``Novel-view x-ray projection synthesis through geometry-integrated deep learning,'' \emph{Med. Image Analy.}, vol.~77, p. 102372, 2022.

\bibitem{Ying2019X2ct}
X.~Ying, H.~Guo, K.~Ma, J.~Wu, Z.~Weng, and Y.~Zheng, ``X2ct-gan: reconstructing ct from biplanar x-rays with generative adversarial networks,'' in \emph{IEEE CVPR}, 2019, pp. 10\,619--10\,628.

\bibitem{Peng2021Xraysyn}
C.~Peng \emph{et~al.}, ``Xraysyn: Realistic view synthesis from a single radiograph through ct priors,'' in \emph{AAAI}, vol.~35, no.~1, 2021, pp. 436--444.

\bibitem{Corona2022Mednerf}
A.~Corona-Figueroa \emph{et~al.}, ``Mednerf: Medical neural radiance fields for reconstructing 3d-aware ct-projections from a single x-ray,'' in \emph{IEEE EMBC}, 2022, pp. 3843--3848.

\bibitem{Cai2024Structure}
Y.~Cai \emph{et~al.}, ``Structure-aware sparse-view x-ray 3d reconstruction,'' in \emph{IEEE CVPR}, 2024, pp. 11\,174--11\,183.

\bibitem{Cai2024Radiative}
Y.~Cai, Y.~Liang \emph{et~al.}, ``Radiative gaussian splatting for efficient x-ray novel view synthesis,'' in \emph{ECCV}, 2024, pp. 283--299.

\bibitem{Suvorov2022Resolution}
R.~Suvorov \emph{et~al.}, ``Resolution-robust large mask inpainting with fourier convolutions,'' in \emph{IEEE WACV}, 2022, pp. 2149--2159.

\bibitem{Sinha2022NL}
A.~K. Sinha \emph{et~al.}, ``{NL-FFC}: Non-local fast fourier convolution for image super resolution,'' in \emph{CVPRW}, 2022, pp. 467--476.

\bibitem{Jaderberg2015Spatial}
M.~Jaderberg \emph{et~al.}, ``Spatial transformer networks,'' \emph{NIPS}, vol.~28, 2015.

\bibitem{Meng2022Artificial}
N.~Meng \emph{et~al.}, ``An artificial intelligence powered platform for auto-analyses of spine alignment irrespective of image quality with prospective validation,'' \emph{eClinicalMedicine}, vol.~43, 2022.

\bibitem{Isola2017Image}
P.~Isola \emph{et~al.}, ``Image-to-image translation with conditional adversarial networks,'' in \emph{IEEE CVPR}, 2017, pp. 1125--1134.

\bibitem{Wang2018High}
T.-C. Wang \emph{et~al.}, ``High-resolution image synthesis and semantic manipulation with conditional gans,'' in \emph{IEEE CVPR}, 2018, pp. 8798--8807.

\end{thebibliography}

\end{document}